# Automatic Lesion Detection System (ALDS) for Skin Cancer Classification Using SVM and Neural Classifiers

Muhammad Ali Farooq
Electronics and Power Engineering (EPE),
Pakistan Navy Engineering College (PNEC), Karachi
National University of Sciences & Technology (NUST), Pakistan
E-Mail: mali.farooq@pnec.nust.edu.pk

Muhammad Aatif Mobeen Azhar
Electronics and Power Engineering (EPE),
Pakistan Navy Engineering College (PNEC), Karachi
National University of Sciences & Technology (NUST), Pakistan
E-Mail: maatif@pnec.nust.edu.pk

Rana Hammad Raza
Electronics and Power Engineering (EPE),
Pakistan Navy Engineering College (PNEC), Karachi
National University of Sciences & Technology (NUST), Pakistan
E-Mail: hammad@pnec.nust.edu.pk

*Abstract* — **Technology aided platforms provide reliable tools in almost every field these days. These tools being supported by computational power are significant for applications that need sensitive and precise data analysis. One such important application in the medical field is Automatic Lesion Detection System (ALDS) for skin cancer classification. Computer aided diagnosis helps physicians and dermatologists to obtain a "second opinion" for proper analysis and treatment of skin cancer. Precise segmentation of the cancerous mole along with surrounding area is essential for proper analysis and diagnosis. This paper is focused towards the development of improved ALDS framework based on probabilistic approach that initially utilizes active contours and watershed merged mask for segmenting out the mole and later SVM and Neural Classifier are applied for the classification of the segmented mole. After lesion segmentation, the selected features are classified to ascertain that whether the case under consideration is melanoma or non-melanoma. The approach is tested for varying datasets and comparative analysis is performed that reflects the effectiveness of the proposed system.**

## I. Introduction

Cancerous mole on skin is a most frequently befalling malignancy in fair skinned populations. There are three basic types of skin cancer, which includes Basal Cell Carcinoma (BCC), Squamous Cell Carcinoma (SCC), and Melanoma. Malignancy is a description of the "stage" of cancer. All of these cancers are fatal however Melanoma comes with the highest risk and found very frequently in the fair skinned people aging less than 50 years for men and more than 50 years for women [17].

During the recent few years, the frequency of melanoma treatment cases has increased extensively, lasting at the top in all the cancers with respect to its management [3]. About 76,380 new cases of melanoma (46,870 for males and 29,510 for females) and about 10,130 cases of new melanoma related life expires are expected during the year 2016 in the United States [11]. Non-melanoma skin cancers, such as Squamous Cell Carcinoma (SCC) or Basal Cell Carcinoma (BCC), contribute to the substantial indispositions among fair skinned Asians [3].

The most important thing from a doctor's point of view is to distinguish and flawlessly identify lesion area. Failure to correctly identify and subsequent delayed treatment of a lesion may lead to advanced stages of cancer. Therefore, early detection is of major importance for the dermatologists. In the process of skin cancer screening, clinicians usually detect the suspected lesion region by visual checkup that is highly dependent on observer skills and is likely to have human error. The authors in [3] have reflected this in their paper asserting that the accuracy of diagnostic abilities can be manipulated as a function of vast diagnostic experience. In European countries, it is highly emphasized to the clinicians to have careful attention and precision in distinguishing and analyzing the skin cancers [3]. Hence by the evolution of improved algorithms and techniques, clinicians may seek "second opinion" from the Automatic Lesion Detection System (ALDS) software to refine their diagnostic performance [3].

The main focus of the proposed research work is to develop a detection and classification framework which can illustrate the non-pigmented skin malignancies along with the pigmented skin growth. The test dataset utilized for this work is the database of Dermatology Service of Hospital, Pedro Hispano (Matosinhos, Portugal) known as PH$^2$ [6]. The PH$^2$ database is based on the process of manual segmentation, the clinical diagnosis, and the identification of several dermoscopic structures which is performed by professional dermatologists. It contains a total of 200 hundred images which include both melanoma and non-melanoma cases. All the images in this dataset are taken under the same conditions through Tuebinger Mole Analyzer system using a magnification of 20x. Other datasets include Dermatology Information System known as DermIS developed by the cooperation of University of Heidelburg (Dept. of Clinical Social Medicine) and University of Erlangen (Dept. of Dermatology) [12], and the dataset of Galderma S.A. known as DermQuest [13]. DermIS.net is the largest dermatology information service available on the internet. It offer elaborate image atlases (DOIA and PeDOIA) complete with diagnoses and differential diagnoses, case reports and additional information on almost all skin diseases including skin cancer. The Paper is focused on combined segmentation approach that utilizes watershed and active contour techniques on the Region of Interest (ROI) by minimizing the energy cost function. The merged mask obtained is then utilized by diagnosis system to map the level of malignancy. The ALDS is motivated by the approach reported by *Chang et al.* [3] from Graduate Institute of Medicine, College of Medicine, Kaoshiung Medical University, Kaohsiung, Taiwan.

## II. BACKGROUND / RELATED WORK

Several computer aided diagnostic systems have been developed based on the modern image processing and computer vision algorithms such as ABCD rule of dermoscopy, Menzies Method,7-Point Checklist , and the CASH algorithm [5]. With the help of these algorithms, the accuracy of the detection and analysis of suspected skin lesion has remarkably improved compared to human visual checkup. By the recent advancements in the image acquisition technology and computer vision algorithms, there has been increased interest in the medical field with an objective to minimize the error and ambiguities in the investigative procedures by yielding a trustworthy "second opinion" for the clinicians. Some of these systems include: Solar Scan [5] designed by the polartechnics Ltd in Australia, DermoGenius-Ultra worked out by LINOS Photonics Inc. DBDermo-MIPS [5] designed by the University of Siena Italy etc. In spite of all these studies, it is worldwide accepted that the accuracy of such systems can be increased no matter to which extent, it is still less. Due to same, this area is under grooming research and requires precise computer adopted algorithms to achieve robust diagnostic processes and tasks [5].

## III. PROPOSED ALGORITHM / APPROACH

In the proposed research work cancerous mole evaluation and classification system has been implemented for the detection of malignant melanoma (skin cancer mole) on any part of body. The proposed algorithm in this paper is as shown in Figure 1.

The algorithm consists of five steps starting from the input phase of preprocessing ranging to the analysis in the form of likelihood of Lesion Malignancy.

*1. Preprocessing:* It includes the process of acquiring the images of the required area. In our case we are using PH² [6] database. All the images have resolution of 765x573 pixels. Moreover we have also used the dataset from DermIS [12] and DermQuest [13]. These datasets contains both melanoma and non-melanoma images of different resolutions. i.e. 550x451, 550x469, 550x367 etc. Images are refined by applying a sharpening filter and hair removal process using dull razor software [8] has been applied for removing clutter.

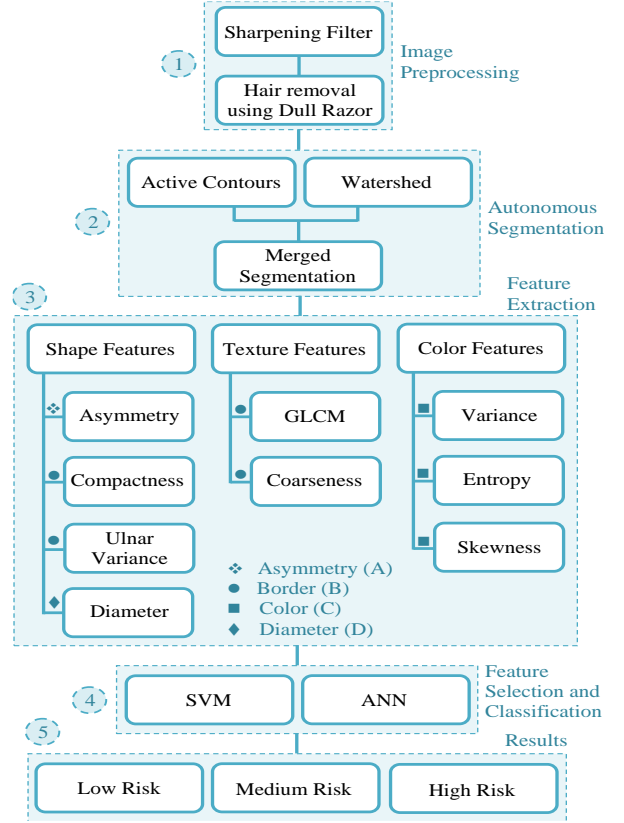

Figure 1: Proposed algorithm of overall system

*2. Segmentation of Cancer Mole:* This phase segments out the cancerous mole autonomously. The segmented region plays an important role to calculate certain features required for further analysis. We have used the merged response of active contour and watershed algorithms for segmentation. Active contours help to learn the point distribution model while watershed helps identify the closed contour. Their individual performance varies and at times goes low in complex cases. The same has been verified through the similarity measures computed in the Section IV. Combining these techniques essentially helps crop the desired region that nearly match the ground truth segmented masks.

*2.1 Watershed Algorithm:* Watershed Algorithm is one of the most efficient and successful region based segmentation techniques used for segmentation especially in the field of medical imaging. As compared to active contours (an iterative process), watershed is less sensitive to noise and is less computationally expensive. Watershed algorithm is a type of topographical representation displaying pixels in the form of valleys and hills where 'valleys' are brighter pixels and 'hills' are the darker pixels [4, 2, 7]. In our case we are using it to segment out the skin cancer mole as it has the ability to efficiently track even weak boundary areas of lesion mole. After extraction, the boundary is smoothed out by means of zero padding and binary image morphological operation so as to obtain final mask of skin cancer mole.

*2.2 Active Contour:* It is a region segmentation technique and is defined as an energy minimizing spline used to locate the object where image structures, such as boundaries exists [10]. It is an iterative process that works to minimize the energy function defined as:

$$E_{Total} = \int_0^1 \left( E_{int}(V(s)) + E_{img}(V(s)) + E_{con}(V(s)) \right) \quad (1)$$

where $E_{int}(V(s))$ is the energy that encourages the prior shape preferences. e.g. Image Smoothness. $E_{img}(V(s))$ is the image energy that forces the contour to fit around the boundaries of the object and $E_{con}(V(s))$ is the energy of contour [1].

*2.3 Merged Segmentation Results:* In our initial findings, the results were not consistent when individual binary masks were used, which led us to merge the masks obtained from active contour and watershed. It is performed by subtracting the watershed segmented image from active contour part which is then complemented and finally the resulting image is multiplied to watershed segmented image. The resulted merged mask then undergoes certain morphological operations like image dilation and image spur for removing spur pixels to get the smoothness and the boundary continuity. As a result, the final mask contains the exact boundary and the necessary surrounding area of the skin lesion, which is not obtained by either mask separately.

*2.4 Similarity Measure:* Once the merged results are obtained, the similarity is computed with reference to the ground truth. This is done by calculating the Structural Similarity Index (SSIM) [14], Jaccard coefficient [15] and Sorenson-Dice coefficient [16] of the computed mask and the ground truth mask provided in the dataset. These coefficients are widely used for the matching purpose and provide better quantitative insight to the change in structural information. The SSIM is defined by the equation.

$$SSIM(x,y) = \frac{(2\mu_x\mu_y + c_1)(2\sigma_{xy} + c_2)}{(\mu_x^2 + \mu_y^2 + c_1)(\sigma_x^2 + \sigma_y^2 + c_2)} \times 100 \quad (2)$$

where $\mu_x, \mu_y, \sigma_x, \sigma_y,$ and $\sigma_{xy}$ are the local means, standard deviations and the cross co-variance of the two image objects $x$ and $y$ and $c_1$ and $c_2$ are some index parameters.

On the other hand, the Jaccard coefficient is a common technique to find the similarity index between the binary variables. It is defined as the intersection and union quotient of the pairwise compared variables between two different binary objects.

$$d^{JAS} = \frac{a}{a+b+c} \times 100 \quad (3)$$

Where $d^{JAS}$ is a symbol of Jaccard coefficient, $a$ is the number of common variables in both objects, b is the number of unique variables in first object and c is the number of unique variables in second object.

Similarly, the Sorenson-Dice coefficient is also a similarity measure approach defined as,

$$d(A,B) = \frac{2|A \cap B|}{|A| + |B|} \times 100 \quad (4)$$

where A and B are the unique variables in object A and B.

*3. Feature Extraction:* This part of algorithm computes the features related to shape, texture and color from the segment mole. Using these three features categories, this approach collects ABCD data of lesion mole where A defines the asymmetry, B corresponds to border Evolving, C is related to color variation and D is the diameter of the mole These features are represented with ❖, ♦,■,● symbols respectively in Figure 1. The shape features includes image diameter, image compactness, ulnar variance and the image asymmetry. Compactness refers to the ratio of the object perimeter to its area. Ulnar variance is the measure of relative length of articular surfaces of some particular radius and the image asymmetry is the measure of asymmetry of the cancerous mole. The texture features include coarseness and Gray Level Co-occurrence Matrix (GLCM). Coarseness is the measure of different angle texture representation. GLCM is a histogram of co-occurring grayscale values for given offset over the image and provides the feature discriminatory attributes. It consists of different parameter which includes mean, correlation, homogeneity, contrast, energy, dissimilarity and kurtosis. The color related features include variance, skewness, and entropy. Variance is the measure of dispersion in the image. Entropy is the proportion of randomness and skewness is the measure of distributed asymmetry.

*4. Feature Selection and Classification:* This phase utilizes supervised learning techniques. i.e. Support Vector Machines (SVM) and the Artificial Neural Networks (ANN). SVM assign weights to all the selected features excluding diameter as it will vary for each test image. A total combination of 73 conventional features were selected from shape, texture and color features group. The first three features were extracted from shape features group which includes asymmetry, image compactness and ulnar variance. From the texture group 46 features were extracted among which 44 were selected from GLCM and 2 from coarseness. Remaining 24 features were selected from color feature group in the form of 8 variance, 8 entropy and 8 skewness features. The feature having smallest weight is considered as the least informative feature and eventually gets rejected. This elimination process iterates till all the outliers are eliminated. Hence, SVM comes up with the best features that are utilized for further mole analysis. Once good features are selected, SVMs is used to differentiate between melanoma and non-melanoma images. Further to improve system performance, ANN classifier [9] is used as a second level classifier to reevaluate the results obtained from SVM and to classify the cases where SVM fails i.e indeterminate cases. The ANN classifier uses eight feature inputs that include mean, correlation, homogeneity, contrast, energy, kurtosis, dissimilarity and skewness. There are ten hidden neurons and one output neuron. Neural Classifier is trained using back propagation algorithm by providing the eight feature values of each case in input matrix and desired outputs in target matrix.

5.  *Results:* The last phase is to classify the segmented mole based on the malignancy probability using SVMs and numbers of cancerous conditions matched using ANN. It is the most important part that helps the physician to diagnosis the case under consideration. SVM provide results in the form of high risk i.e. melanoma, low risk i.e. non-melanoma or medium risk i.e. indeterminate. Indeterminate case can be further diagnosed by using ANN classifier or performing biopsy test. The Probability for SVMs as given in [3] can be calculated as,

$$P = 2 * (F - min) / (max - min) - 1 \quad (5)$$

(Where F = Image feature vector, min = Minimum feature values of cancerous area, max = Maximum feature values).

## IV. IMPLEMENTATION AND RESULTS

The proposed algorithm has been implemented using MATLAB R2013B ® and tested on the Intel Core i5 based computing platform. The whole process of classification takes approximately up to 4 minutes. The test was performed on 50 cases among which certain cases were selected for proposed research work. The selected images and their corresponding masks from the dataset [6, 12-13] are shown in Figure 2. The implementation consists of two steps:

1.  *Binary Mask Computation:* This step consists of the segmentation task, in which the cancerous area is segmented. The watershed algorithm provides the segmentation results as shown in Figure 7. The masks computed for all the test images using watershed, active contour and merging both techniques along with the ground truth are shown in Figure 2. The overall all process of watershed and active contour merged approach is separately shown in Figure 6.

The similarity measure is calculated by comparing each computed mask with the ground truth mask. The results of Structural Similarity Index (SSIM), Jaccard's coefficient and Sorenson-Dice coefficient are provided in Table II.

It can be observed that the similarity measure of the merged mask with ground truth mask is maximum. It is to be noted that the available datasets have images taken in controlled illumination and pose. Any variation in these parameters is expected to affect these results. Such an example is shown in Figure 3.

2.  *Evaluation of Cancerous Mole:* Once the lesion mole has been segmented out the mask is loaded on actual rgb image shown in Figure 4. The next step is to evaluate the mole. SVM classify the result in form of either Low risk i.e. non melanoma, Medium risk i.e. indeterminate and High risk i.e. melanoma. This is done by computing the necessary features and the probability percentage (eq. 5) as discussed in Section III.

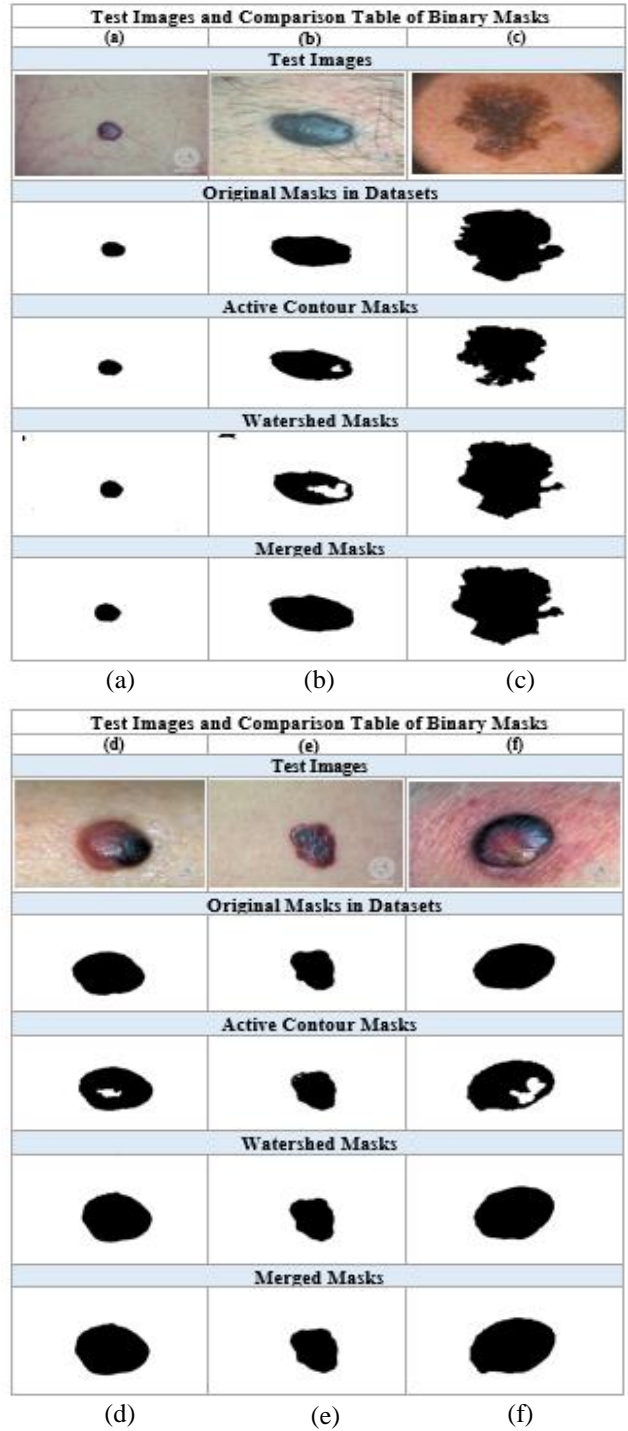

Figure 2: Test images and comparison of binary masks   (Dataset [6, 12-13])

However, SVM may provide varying results depending upon the skewed or imbalanced datasets and often classify the case in the form of medium level risk i.e. Indeterminate Case Therefore, Artificial Neural Networks (ANN) are also used in a complementary role. ANN classifies the mole in the form of zero (Low risk) and one (High risk) where one indicates melanoma (cancerous) mole and zero indicates non melanoma mole. The ANN classifier is trained using both melanoma and non-melanoma cases. The results contain six misclassification and accuracy of the system is 80% as compared to dermatologist's findings. The neural network confusion matrix and ROC are shown Figure 5.

The classification rule for SVMs and for Neural Classifier is provided in Table I. The classification results based on these rule are shown in Table III which shows that the dataset images of Figure 2 (a) and (b) have the low probability (i.e. < 50%) of becoming melanoma and hence are classified as low risk level. Similarly, the image of Figure 2 (c), (d) and (f) has the greater probability (i.e. ≥ 50%) hence, it is classified as high risk level which means that it is a melanoma case. Whereas Figure 2 (e) have medium level risk using SVM which was further classified using ANN and diagnosed as high level risk. These results also matches the ground truth provided within databases.

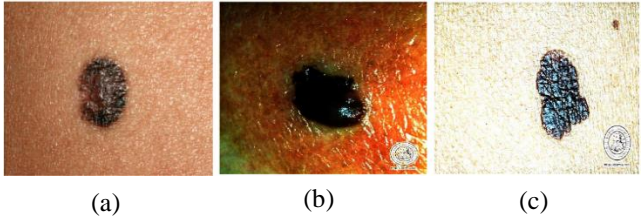

(a)       (b)       (c)

Figure 3: Images under changed lighting conditions.
(a). Good lighting condition. (b) & (c) Inappropriate lighting

TABLE I: CLASSIFICATION RULES

| For SVM | |
|---|---|
| Probability > 0.40 | Risk Level: High |
| Probability > 0.30 | Risk Level: Medium |
| Probability < 0.30 | Risk Level: Low |
| For Neural Classifier | |
| Matching Probability ≥ 50% | Risk Level: High (Output 1) |
| Matching Probability < 50% | Risk Level: Low (Output 0) |

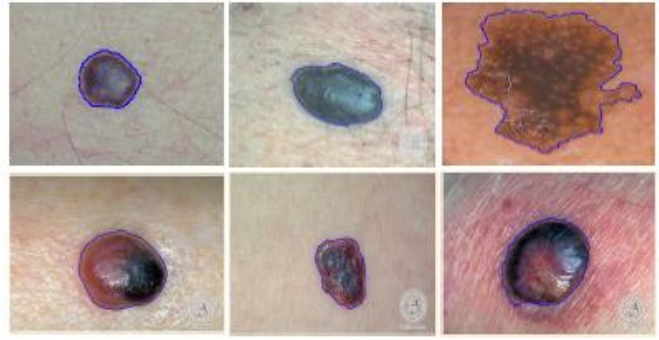

Figure 4: Six different cases of different categories with masks loaded on actual images.

TABLE II: SIMILARITY MEASURE

| Images | Active Contour Mask & Original Mask | Watershed Mask & Original Mask | Merged / Intersected Mask & Original Mask |
|---|---|---|---|
| Structural Similarity Index (%) | | | |
| Fig. 2 (a) | 99.7 | 99.6 | 99.8 |
| Fig. 2 (b) | 99.8 | 99.6 | 99.9 |
| Fig. 2 (c) | 98.9 | 99.1 | 99.6 |
| Fig. 2 (d) | 99.8 | 99.8 | 99.9 |
| Fig. 2 (e) | 99.95 | 99.92 | 99.96 |
| Fig. 2 (f) | 99.37 | 99.65 | 99.72 |
| Jaccard Coefficient (%) | | | |
| Fig. 2 (a) | 99.7 | 99.7 | 99.85 |
| Fig. 2 (b) | 98.3 | 96.5 | 99.16 |
| Fig. 2 (c) | 89.6 | 93 | 95.8 |
| Fig. 2 (d) | 98.3 | 98.5 | 99.1 |
| Fig. 2 (e) | 99.5 | 99.3 | 99.6 |
| Fig. 2 (f) | 94.48 | 96.7 | 97.42 |
| Sorenson Dice Coefficient (%) | | | |
| Fig. 2 (a) | 99 | 99.2 | 99.85 |
| Fig. 2 (b) | 99 | 98.2 | 99.6 |
| Fig. 2 (c) | 94.5 | 96 | 97.9 |
| Fig. 2 (d) | 99.1 | 99.3 | 99.5 |
| Fig. 2 (e) | 99.7 | 99.6 | 99.8 |
| Fig. 2 (f) | 97.16 | 98.14 | 98.69 |

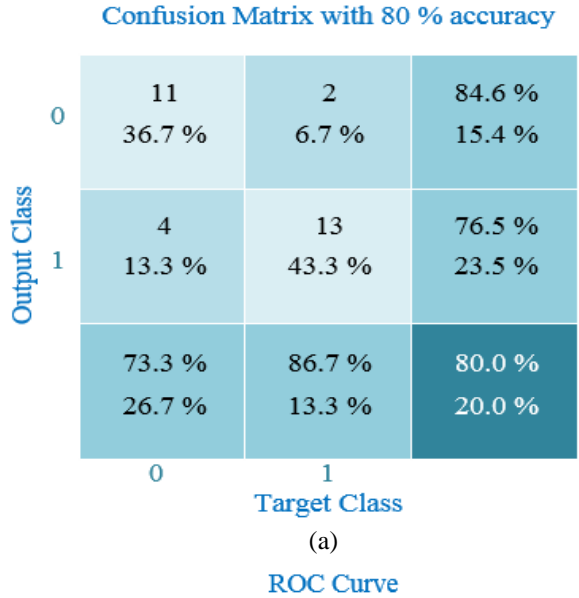

(a)

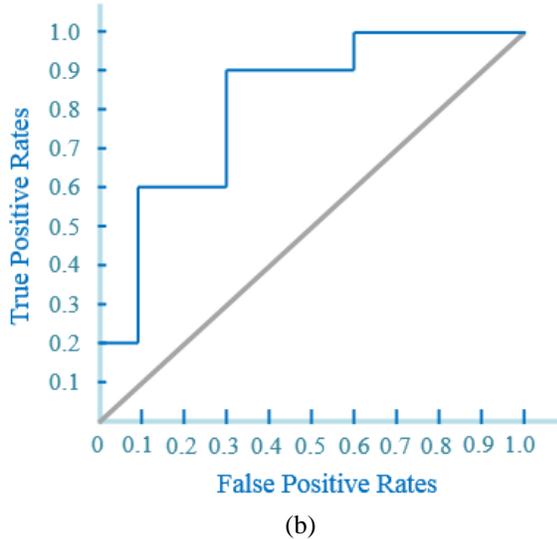

(b)

Figure 5: Neural Classifier Results (Confusion Matrix and Receiver Operative Curve)

CONCLUSION / FUTURE WORK

The Automatic Lesion Detection System (ALDS) for skin cancer classification is the extended work of *Chang et al.* [3]. Initially sharpening filter is applied and also hair removal is performed using dull razor software [8] that eventually produces more refined results. Active contours and watershed approaches are used to segment out the cancerous area automatically from the dataset image with increased efficiency, whereas the classification of cancer mole using SVM was practiced using research findings of Chang et al [3]. On the other hand ANN classifier is implemented as second level classifier to witness the results obtained from SVM and also to check the cases where SVM fails to classify i.e. indeterminate cases. The proposed research work concludes that merged approach of watershed and active contour produces better segmented output.

Subsequently SVM and ANN complement each other and help to provides better classification decisions. Due to fewer number of freely available datasets and having small number of image samples (in controlled environment) with subsequent medical diagnosis, there are certain aspects which cannot be tested at this stage and needs further analysis as future work. These include more than one cancerous mole in an image, varying illumination conditions and slightly varying pose of the region of interest. These factors are expected to change the features characteristics and is an open problem to research community. Moreover the accuracy of neural networks can also be further enhanced by increasing the number of samples and the number of features in the training phase. Additionally, in future this system can be made to learn the evolution of the cancerous mole before time based on probabilistic and forecasting measures. This will require test images of the same case progressing over time.

TABLE III: CLASSIFICATION RESULTS

| SVMs Results | | | |
|---|---|---|---|
| Image | Probability (%age) with Original Mask | Probability (%age) with *Merged Mask | Classification |
| Fig. 2 (a) | 16.2 | 19.4 | Risk Level: Low |
| Fig. 2 (b) | 18.7 | 21.7 | Risk Level: Low |
| Fig. 2 (c) | 58.4 | 51.5 | Risk Level: High |
| Fig. 2 (d) | 40.5 | 46.5 | Risk Level: High |
| Fig. 2 (e) | 33.1 | 32.3 | Risk Level: Medium |
| Fig. 2 (f) | 77.0 | 75.8 | Risk Level: High |
| Neural Classifier Results | | | |
| Image | Classification | | |
| Figure 2 (a) | 0 (Risk Level: Low) | | |
| Figure 2 (b) | 0 (Risk Level: Low) | | |
| Figure 2 (c) | 1 (Risk Level: High) | | |
| Figure 2 (d) | 1 (Risk Level: High) | | |
| Figure 2 (e) | 1 (Risk Level: High) | | |
| Figure 2 (f) | 1 (Risk Level: High) | | |

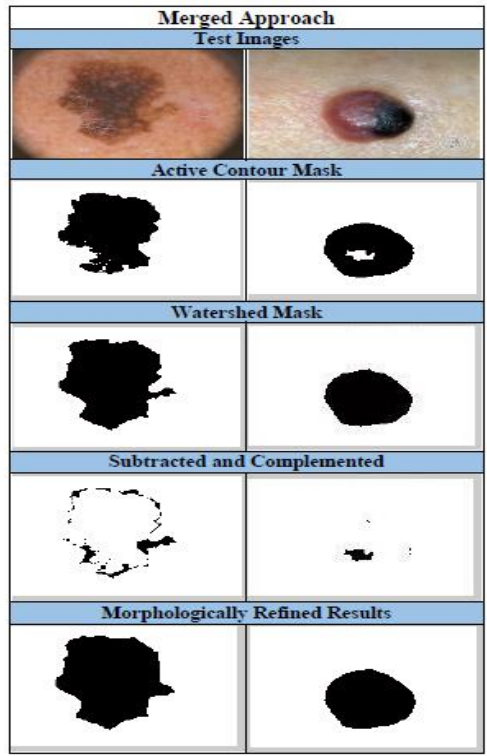

Figure 6: Watershed and Active contour merged process

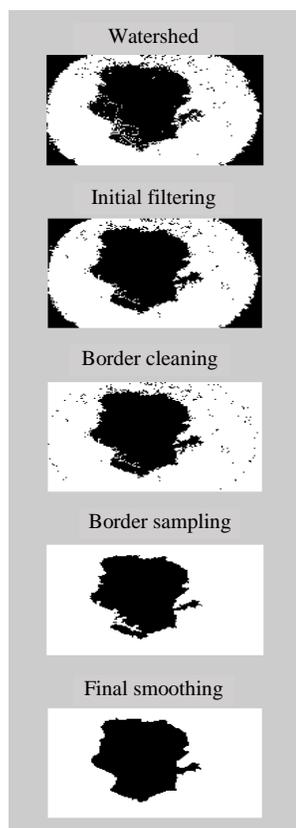

Figure 7: Watershed algorithm detailed process

ACKNOWLEDGMENT

The authors would like to acknowledge the University of Heidelburg and University of Erlangen the contributors of DermIS dataset, Galderma S.A for DermQuest and PH$^2$ database contributors for providing the image resources to carry out this research work.